\documentclass[journal]{IEEEtran}

\usepackage[T1]{fontenc}
\usepackage[utf8]{inputenc}
\usepackage{graphicx}
\graphicspath{{figures/}}
\usepackage{amsmath,amssymb,amsfonts}
\usepackage{amsthm}
\usepackage{mathrsfs}
\usepackage{textcomp}
\usepackage{xcolor}
\usepackage{booktabs}
\usepackage{multirow}
\usepackage{makecell}
\usepackage{cite}
\usepackage[hidelinks]{hyperref}
\usepackage{url}
\usepackage{caption}
\usepackage{subcaption}
\usepackage{placeins}

\usepackage{url}
\begin{document}
\title{Benchmarking State Space Models, Transformers, and Recurrent Networks for US Grid Forecasting}

\author{Sunki~Hong and Jisoo~Lee,~\IEEEmembership{Member,~IEEE}
\thanks{S. Hong and J. Lee are with Gramm AI (e-mail: sunki@gramm.ai; jisoo@gramm.ai).} \thanks{Corresponding author: Jisoo Lee.}}

\maketitle
\begin{abstract}
Selecting the right deep learning model for power grid forecasting is challenging, as performance heavily depends on the data available to the operator. This paper presents a comprehensive benchmark of five modern neural architectures---two state space models (PowerMamba, S-Mamba), two Transformers (iTransformer, PatchTST), and a traditional LSTM. We evaluate these models on hourly electricity demand across six diverse US power grids for forecast windows between 24 and 168 hours. To ensure a fair comparison, we adapt each model with specialized temporal processing and a modular layer that cleanly integrates weather covariates.

Our results reveal that there is no single best model for all situations. When forecasting using only historical load, PatchTST and the state space models provide the highest accuracy. However, when explicit weather data is added to the inputs, the rankings reverse: iTransformer improves its accuracy three times more efficiently than PatchTST. By controlling for model size, we confirm that this advantage stems from the architecture's inherent ability to mix information across different variables. Extending our evaluation to solar generation, wind power, and wholesale prices further demonstrates that model rankings depend on the forecast task: PatchTST excels on highly rhythmic signals like solar, while state space models are better suited for the chaotic fluctuations of wind and price. Ultimately, this benchmark provides grid operators with actionable guidelines for selecting the optimal forecasting architecture based on their specific data environments.
\end{abstract}


\section{Introduction}

\IEEEPARstart{S}{hort-term} load forecasting (STLF) underpins electricity grid operations, governing unit commitment and economic dispatch \cite{hong2016probabilistic,ahmad2024short}. Despite rapid architectural innovation---from LSTMs to Transformers to state space models (SSMs) \cite{gu2022efficiently}---most deep learning forecasting studies evaluate on a handful of grids (typically 1--5). This limited scope makes it difficult to assess whether architectural advantages generalize across the diverse characteristics of US grid operators \cite{dong2024comprehensive}.

This limitation is particularly pressing as US grids face increasing complexity. Behind-the-meter solar creates steep net-load ramps (the ``duck curve'') in CAISO; extreme weather can trigger non-linear demand spikes in ERCOT; and each ISO exhibits distinct load profiles shaped by regional climate, industrial mix, and renewable penetration. Three model families currently compete---recurrent networks (LSTMs), Transformer variants, and state space models---but whether their advantages hold across US grid diversity and interact with weather covariate availability remains an open question. Practitioners need reliable guidance: \textit{which architecture should they deploy for their specific grid?}

We address this gap with a systematic benchmark of five neural architectures across six major US independent system operators. Using hourly load data from EIA-930, we evaluate each architecture under identical preprocessing, training, and evaluation protocols across forecast windows from 24 to 168 hours, with additional analysis of weather feature integration.

\begin{enumerate}
    \item \textit{US Grid Benchmark.} We conduct a comprehensive comparison of SSMs (PowerMamba, S-Mamba), Transformers (iTransformer, PatchTST), and an LSTM across six major US ISOs with consistent hyperparameters. PatchTST leads by outperforming on 15 of 30 evaluations (MAPE), followed by SSMs (14 of 30 combined). We report MSE (\%), MAPE (\%), and signed-error tails ($P_{0.5}/P_{99.5}$) in an operator-by-window layout (Section~\ref{sec:results}).

    \item \textit{Weather Integration Benchmark.} We develop thermal-lag-aligned weather fusion strategies for each architecture and evaluate all five models across seven US grids at $W=24$. iTransformer's cross-variate attention achieves $3\times$ greater weather benefit than PatchTST's channel-independent design ($-1.62$ vs.\ $-0.52$ percentage points average $\Delta$MAPE), revealing that weather integration effectiveness is architecturally determined (Section~\ref{sec:weather}).

    \item \textit{Multitype Generalization.} We extend the benchmark to solar generation, wind generation, wholesale price, and ancillary-service forecasting across US grids, showing that architectural rankings are task-dependent: PatchTST dominates intermittent signals while SSMs lead smoother signals (Section~\ref{sec:multitype}).
\end{enumerate}

\section{Background and Related Work}

\subsection{Operational Role of Short-Term Load Forecasting}
Short-term load forecasting (STLF) predicts aggregate electricity demand from hours to days ahead and is a core input to system operations \cite{hong2016probabilistic,ahmad2024short,dong2024short}. US wholesale markets operate under a two-settlement structure---day-ahead clearing based on forecasted demand and real-time settlement of deviations \cite{wood2014power}---with STLF feeding both stages. Security-constrained unit commitment (SCUC) and economic dispatch (SCED) rely directly on load forecasts, and operating reserve requirements scale with expected forecast error \cite{nerc2023balancing}. Our benchmark windows (Section~\ref{sec:task_definition}) span the operational range from real-time dispatch (1--6~h) through day-ahead commitment (12--36~h) to resource adequacy planning (48--168~h).

\subsection{Deep Learning for Time Series Forecasting}
Deep learning has largely displaced statistical methods (ARIMA, SVR) for short-term load forecasting \cite{dong2024comprehensive,eren2024systematic}. The power systems community has actively embraced deep neural networks to manage the increasing stochasticity of smart grids. Recent studies in premier venues like the \textit{IEEE Transactions on Smart Grid} and \textit{IEEE Transactions on Power Systems} have demonstrated the efficacy of deep learning across diverse forecasting challenges. Innovations include deep residual networks \cite{Chen_2019}, densely connected networks \cite{Li_2021}, pooling-based deep RNNs \cite{Shi_2018}, and ensemble learning frameworks \cite{Su_2024,Von_Krannichfeldt_2021} for deterministic structural forecasting. To capture uncertainty, researchers have proposed Bayesian deep learning \cite{Sun_2020}, probabilistic multitask prediction models \cite{Wang_2024}, transfer-learning frameworks for masked behind-the-meter loads \cite{Wu_2023,Zhou_2022,He_2022}, collaborative multi-district approaches \cite{Liu_2024}, and advanced quantile-mixing methods \cite{Ryu_2024}.

While these studies have successfully introduced powerful bespoke architectures tailored to specific challenges (e.g., behind-the-meter solar, probabilistic limits), they often evaluate models on restricted datasets or a few isolated operators. Our work complements this rich literature by providing a systematic, cross-architecture benchmark---evaluating pure SSMs, Transformers, and RNN paradigms on identical footing across six diverse US market operators and systematically quantifying how weather covariate integration interacts with each foundational architecture. Specifically, within the Transformer family, two multivariate strategies have emerged: channel-independent patching \cite{nie2023time} and cross-variate tokenization \cite{liu2023itransformer}. In parallel, state space models have brought linear-time complexity to sequence modeling \cite{gu2022efficiently,gu2023mamba}, challenging the quadratic scaling of Transformers. LSTMs remain the most widely deployed class in operational settings \cite{hochreiter1997long,bouktif2020optimal}. We benchmark these pure architectural paradigms to establish their foundational behavior before they are blurred by hybrid designs (e.g., Mambaformer).

While zero-shot foundation models (TimesFM \cite{das2024timesfm}, Chronos \cite{ansari2024chronos}) are emerging, this paper focuses on trained-from-scratch (TFS) architectures that can natively integrate grid-specific, delay-aligned weather covariates---a capability that current zero-shot models lack without extensive fine-tuning.

Each paradigm has been evaluated on energy data, but rarely against each other under identical conditions. PowerMamba \cite{menati2024powermamba} benchmarked a bidirectional SSM on five US grids; S-Mamba \cite{wang2024effective} demonstrated SSM effectiveness on general multivariate benchmarks but was not evaluated on US grid data; PatchTST \cite{nie2023time} established records on ETTh/ETTm datasets with limited grid diversity. Direct comparisons between LSTMs and modern SSMs under production-like conditions are scarce, and systematic cross-architecture evaluation of weather integration remains limited \cite{hong2016probabilistic,seem2007dynamic,eren2024systematic}. No prior work benchmarks SSM, Transformer, and RNN architectures across multiple US ISOs under identical protocols. Our benchmark addresses these gaps: five models across six US grids (seven for weather experiments), with architecture-matched weather integration and parameter-controlled comparisons to disentangle capacity from design.

\section{Problem Formulation and Data Characteristics}

\subsection{Forecast Task Definition}
\label{sec:task_definition}
We study \textit{system-level} (balancing-authority aggregate) load forecasting rather than nodal or zonal forecasting. System-level forecasting is the primary input to SCUC/SCED and is the level at which ISOs report forecast accuracy. We use EIA-930 hourly demand series \cite{eia930}, the only publicly available dataset providing consistent, hourly, multi-ISO load data across the US---making it uniquely suited for cross-operator benchmarking.

Given a historical window of length $L$, the model predicts the next $W$ steps:
\begin{equation}
\hat{\mathbf{y}}_{t+1:t+W} = f_{\theta}\!\left(\mathbf{x}_{t-L+1:t}, \mathbf{z}_{t-L+1:t}\right),
\end{equation}
where $\mathbf{x}$ is historical load and $\mathbf{z}$ denotes optional exogenous features (temporal embeddings and, when enabled, weather covariates). This formulation supports two deployment-relevant settings: load-only inference ($\mathbf{z}$ = calendar context) and weather-augmented inference ($\mathbf{z}$ includes meteorological covariates). Comparing both settings clarifies when architectural advantages depend on exogenous information availability. Specific values of $L$ and $W$ are defined in Section~\ref{sec:dataset}.

\subsection{Signal Characteristics and Exogenous Drivers}
\label{sec:signal_characteristics}
Grid load signals are multi-periodic and non-stationary. Strong daily (24-hour) and weekly (168-hour) cycles dominate, with typical peak-to-trough ratios of 1.3--1.7$\times$ within a day and weekend loads 5--15\% below weekday levels. These periodicities coexist with trend shifts from weather, holidays, and economic activity.

ISO-specific characteristics shape the forecasting challenge. CAISO exhibits the ``duck curve''---midday net-load troughs from 30\%+ solar penetration followed by steep evening ramps. ERCOT is electrically isolated from the Eastern and Western Interconnections, amplifying local weather sensitivity and reducing spatial load diversity. PJM and MISO, as the two largest US interconnections, benefit from geographic smoothing across diverse climate zones. ISO-NE is winter-peaking due to electric heating prevalence, unlike the summer-peaking pattern of most US grids. Non-stationarity is accelerating as behind-the-meter solar growth progressively reduces observable load during daylight hours, creating a divergence between gross and net demand that complicates model training on historical data.

Weather effects are mediated by a U-shaped temperature--load relationship: heating demand rises below ${\sim}18$\textdegree{}C and cooling demand rises above it, with a comfort zone of minimal weather sensitivity near the inflection point \cite{hong2016probabilistic}. The response is delayed by building thermal inertia---HVAC systems respond to temperature changes with 2--6 hour lags depending on building mass \cite{seem2007dynamic}. Solar irradiance and wind speed influence net demand more rapidly but with high intermittency. This motivates the lag-aware weather feature construction used in our weather experiments (Section~\ref{sec:weather}).

\section{Evaluated Architectures}
\label{sec:methodology}

\subsection{Model Architectures}

We evaluate five models spanning three neural architecture families: two state space models representing different SSM design philosophies, two Transformer variants with contrasting multivariate strategies, and an LSTM.

\subsubsection*{S-Mamba}
S-Mamba \cite{wang2024effective} represents a minimalist approach to state space modeling. It adapts the standard Mamba block for time series via a simple encoder-decoder structure: a linear projection embeds the input sequence into a hidden state $D$, followed by stacked Mamba layers.

S-Mamba tests whether the core selective state space mechanism alone---without complex decomposition or attention---suffices for grid forecasting. Its design prioritizes computational efficiency ($O(n)$ complexity), making it a candidate for resource-constrained edge deployment.

\subsubsection*{PowerMamba}
PowerMamba \cite{menati2024powermamba} addresses a specific limitation of standard SSMs: the difficulty of modeling disparate frequencies (e.g., long-term trend vs. daily seasonality) within a single state vector.

PowerMamba introduces two critical innovations for energy data. First, \textit{Series Decomposition} splits the input into "Trend" (low-frequency) and "Seasonal" (high-frequency) components, processing them via independent Mamba encoders. This explicitly separates the duck curve drift from daily load cycles. Second, \textit{Bidirectionality} captures a holistic temporal context; unlike language modeling where causality is strict, time series analysis benefits from looking forward (e.g., smoothing), so PowerMamba processes sequences in both forward and backward directions.

\subsubsection*{PatchTST}
PatchTST \cite{nie2023time} represents the channel-independent paradigm for Transformer-based time series forecasting. Rather than tokenizing variables, PatchTST segments each univariate channel into fixed-length patches and processes them with a shared Transformer encoder. Channel independence means each variate shares the same Transformer weights, enabling efficient multi-channel processing without explicit cross-variable attention. This tests whether temporal pattern extraction alone suffices for accurate grid forecasting.

\subsubsection*{iTransformer}
The iTransformer \cite{liu2023itransformer} challenges the standard Transformer paradigm by tokenizing \textit{variables} (embedding an entire time series variate as a single token) instead of time steps. Self-attention therefore computes correlations \textit{between variables} (e.g., Load vs.\ Temperature), explicitly modeling system-wide interdependencies. By design, this mechanism requires multiple variates; with a single load variate, self-attention reduces to near-identity.

\subsubsection*{LSTM}
We use a 2-layer bidirectional LSTM \cite{hochreiter1997long} with hidden dimension $d=256$ as our recurrent baseline, yielding approximately 2.6M parameters at $W=48$.

\subsection{Architecture Adaptations for Grid Forecasting}
\label{sec:adaptations}

Direct application of generic time series models to load forecasting yields suboptimal results because these architectures lack two capabilities essential for grid operations: (1) awareness of \textit{temporal context}---hourly and weekly demand cycles that dominate load profiles---and (2) the ability to \textit{selectively incorporate weather covariates} whose influence on load is non-stationary and lagged.

We adapt each baseline with a shared set of domain-specific components while preserving the core architectural mechanism that each model is designed to test:

\textit{Temporal embeddings.} All sequence models (SSMs, LSTM) receive learnable hour-of-day ($\mathbb{R}^{24 \times d/4}$) and day-of-week ($\mathbb{R}^{7 \times d/4}$) embeddings, concatenated and projected to $d_{\text{model}}$, then added to the input representation at each time step. PatchTST processes temporal context implicitly through its patch structure; iTransformer encodes hour and day as additional variate tokens. These embeddings inject the strong 24h and 168h periodicities that characterize load without requiring the encoder to learn them from data alone.

\textit{Bidirectional encoding.} SSMs and LSTM use bidirectional processing---separate forward and backward encoders whose outputs are fused via a learned linear projection. Crucially, bidirectional processing is applied \textit{strictly within the fixed-length lookback window} ($L=240$ hours); no information from the forecast horizon $W$ is accessible to the encoder. Unlike causal language models, load forecasting operates on a completed historical window where ``future'' context within that window is available and informative. Bidirectional encoding approximately doubles the encoder's parameter count relative to a unidirectional model, but captures both leading and lagging temporal relationships (e.g., a morning ramp informs the preceding overnight trough and vice versa).

\textit{Attention-weighted pooling.} Rather than mean-pooling the encoder's hidden states, SSMs and LSTM use a learned attention vector ($\mathbf{a} = \text{softmax}(\mathbf{W}_a \mathbf{H})$) to produce a weighted summary $\mathbf{c} = \sum_t a_t \mathbf{h}_t$. This allows the model to selectively emphasize time steps most relevant to the forecast (e.g., weighting recent ramp events more heavily than stable overnight periods).

\textit{Modular weather integration.} Each architecture receives weather covariates through a fusion mechanism matched to its inductive bias (detailed in Section~\ref{sec:weather_arch}). Critically, the weather integration layer is architecturally \textit{decoupled} from the core encoder: it can be toggled on or off by providing or withholding weather features at inference time, without retraining. This modularity enables the controlled experiments in Section~\ref{sec:weather} and the fair weather comparison in Section~\ref{sec:fair_comparison}.

\textit{Impact on parameter counts.} These adaptations increase parameter counts relative to original configurations (e.g., bidirectional encoding doubles SSM encoder parameters; weather cross-attention adds $\sim$330K per model). All counts reported reflect the full deployed configuration at $W=48$; see the Supplementary Material for a per-model parameter breakdown.

\subsection{Weather-Integrated Architectures}
\label{sec:weather_arch}

\begin{figure*}[htbp]
  \centering
  \begin{subfigure}[t]{0.49\textwidth}
    \centering
    \includegraphics[width=\linewidth]{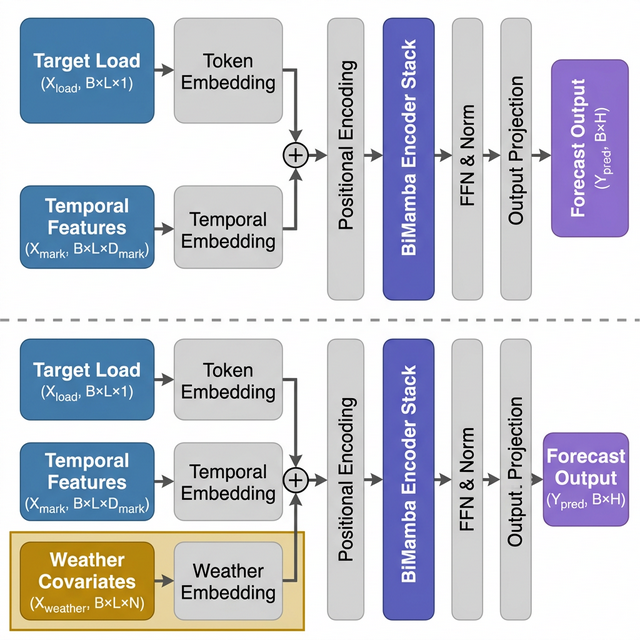}
    \caption{S-Mamba: Early Fusion via Summation}
    \label{fig:smamba}
  \end{subfigure}
  \hfill
  \begin{subfigure}[t]{0.49\textwidth}
    \centering
    \includegraphics[width=\linewidth]{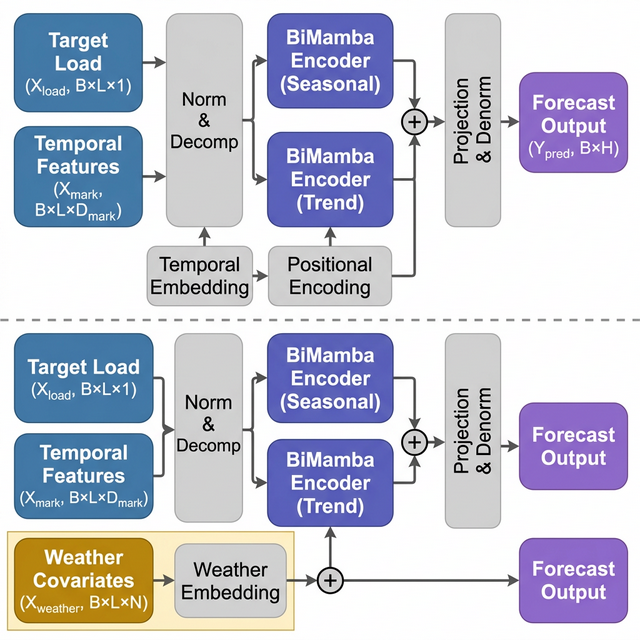}
    \caption{PowerMamba: Summation into decomposed streams}
    \label{fig:powermamba}
  \end{subfigure}

  \vspace{0.5em}

  \begin{subfigure}[t]{0.49\textwidth}
    \centering
    \includegraphics[width=\linewidth]{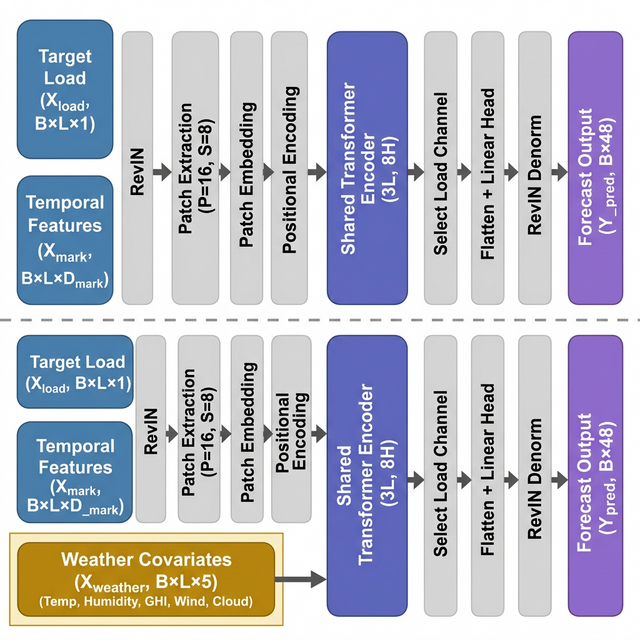}
    \caption{PatchTST: Channel Fusion via Independent Patching}
    \label{fig:patchtst}
  \end{subfigure}
  \hfill
  \begin{subfigure}[t]{0.49\textwidth}
    \centering
    \includegraphics[width=\linewidth]{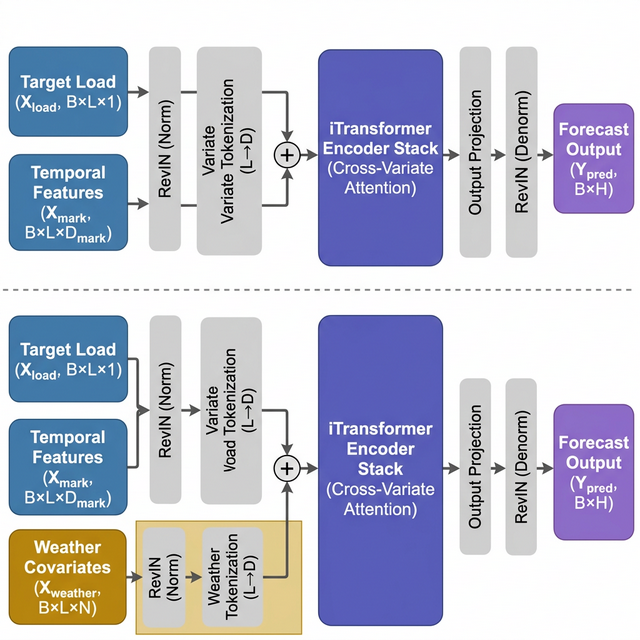}
    \caption{iTransformer: Weather as additional tokens}
    \label{fig:itransformer}
  \end{subfigure}
  \caption{\textbf{Weather integration strategies.} Each subfigure shows the baseline architecture (top) and its weather-integrated variant (bottom). (a) S-Mamba: early fusion by embedding-space summation. (b) PowerMamba: summation into decomposed streams. (c) PatchTST: channel fusion via independent patching. (d) iTransformer: tokenization of weather variables for cross-variate attention.}
  \label{fig:architectures}
\end{figure*}

The key challenge in weather fusion is respecting the building thermal lag described in Section~\ref{sec:signal_characteristics}. We developed weather-integrated variants of each architecture (Figure~\ref{fig:architectures}), all following an \textit{encode-then-fuse} pattern: the core encoder first processes load history and temporal features, then a secondary fusion layer incorporates meteorological context. This design ensures that the same encoder weights serve both load-only and weather-augmented inference, and that weather integration adds a fixed parameter overhead ($\sim$0.3M for SSMs/LSTM, $\sim$0.7M for PatchTST), enabling fair cross-architecture comparison.

\subsubsection*{Fusion Strategies (Architecture-Matched)}

\textit{S-Mamba (Early Summation, Fig.~\ref{fig:smamba}).} By summing weather embeddings with load and temporal tokens prior to the BiMamba stack, all exogenous signals are projected into a unified $d_{\text{model}}$ dimensional space. The selective state space mechanism then uses this fused representation to modulate its continuous transition matrices ($\mathbf{A}, \mathbf{B}, \mathbf{C}$), allowing weather to directly gate the recurrent state updates at each timestep.

\textit{PowerMamba (Pre-Decomposition Fusion, Fig.~\ref{fig:powermamba}).} Weather features are fused before the sequence is decomposed. Because meteorological events (e.g., prolonged heatwaves) drive both multi-day trend shifts and rapid sub-daily HVAC cycling, early fusion allows the moving-average decomposition module to appropriately route weather-induced variance into both the seasonal and trend BiMamba branches.

\textit{PatchTST (Interleaved Cross-Attention, Fig.~\ref{fig:patchtst}).} To preserve PatchTST's core channel-independence inductive bias---which prevents direct mixing of variates to avoid overfitting---we patch load and weather channels independently. We then insert a dedicated cross-attention sublayer where load patches query the weather patches. This permits the load representation to attend to relevant meteorological context (e.g., a temperature spike) without entangling the self-attention dynamics of the load and weather streams.

\textit{iTransformer (Variate Tokenization, Fig.~\ref{fig:itransformer}).} iTransformer is natively designed to model cross-variate correlations. Therefore, each weather covariate is embedded as a distinct variate token sequence mapping the entire lookback window. These weather tokens are concatenated with load and temporal tokens, exposing them directly to the global cross-variate attention mechanism and enabling dynamic, pairwise correlation modeling (e.g., Load heavily attending to Temperature during summer peaks).

\textit{LSTM (Early Concatenation).} Given the step-by-step recurrent processing of the LSTM, weather features are concatenated with load features at each timestep. This ensures the recurrent gating mechanisms (input, forget, and output gates) have simultaneous access to both the current demand and contemporaneous meteorological forcing, allowing the hidden cell state to accumulate weather-driven momentum over time.

\section{Experiments and Results}

\subsection{Experimental Setup}

\subsubsection*{Dataset and Splits}
\label{sec:dataset}
We use hourly system-level load data from six major US independent system operators, sourced from the EIA-930 Hourly Electric Grid Monitor \cite{eia930}. Each ISO provides approximately four years of hourly load data spanning 2022--2025. Table~\ref{tab:dataset} summarizes the grids.

\begin{table}[htbp]
\centering
\caption{\textbf{US grid operators in the benchmark.} Peak load and primary operating characteristics. $^\dagger$SWPP is used in weather experiments only (Section~\ref{sec:weather}).}
\label{tab:dataset}
\begin{tabular}{llrl}
\toprule
\textbf{ISO} & \textbf{EIA ID} & \textbf{Peak (GW)} & \textbf{Characteristics} \\
\midrule
CAISO & EIA-CISO & 47.1 & High solar, duck curve \\
ISO-NE & EIA-ISNE & 25.6 & Weather-sensitive, winter peaks \\
MISO & EIA-MISO & 127.1 & Large, wind-rich \\
PJM & EIA-PJM & 150.4 & Largest US ISO \\
SWPP & EIA-SWPP & 54.4 & Wind-rich, central US$^\dagger$ \\
ERCOT & EIA-ERCO & 85.5 & Isolated grid, extreme weather \\
NYISO & EIA-NYIS & 32.1 & Dense urban, summer peaks \\
\bottomrule
\end{tabular}
\end{table}

The six core ISOs span the operating characteristics described in Section~\ref{sec:signal_characteristics}; SWPP is added for the weather-extension experiments.

Context length is $L=240$ hours (10 days), following prior evidence that long context windows improve grid forecasting \cite{menati2024powermamba}. The main benchmark evaluates prediction windows $W=\{24,48,72,96,168\}$. For weather integration analysis (Section~\ref{sec:weather}), we evaluate at $W=24$ using five meteorological covariates (temperature, humidity, wind speed, GHI, and cloud cover; full list in the Supplementary Material) aligned with empirically derived thermal lag parameters.

\textbf{Preprocessing.} Load values are z-score normalized per grid using training-set statistics only (no information leakage). Temporal features (hour-of-day, day-of-week) are encoded as learnable embeddings (Section~\ref{sec:adaptations}).

\subsubsection*{Model Set and Training Protocol}
\label{sec:training}
We benchmark five models from three neural architecture families: S-Mamba, PowerMamba, PatchTST, iTransformer, and LSTM. Architecture definitions and grid-specific adaptations are described in Section~\ref{sec:adaptations} and Section~\ref{sec:weather_arch}.

Core model scales are fixed across operators for comparability ($d_{\text{model}}=256$ for SSMs/PatchTST, $d_{\text{model}}=512$ for iTransformer, and 2-layer bidirectional LSTM). Full architecture-by-architecture settings are reported in the Supplementary Material.

All models are trained with AdamW optimizer (lr $10^{-3}$, weight decay $10^{-4}$), OneCycleLR scheduler, up to 200 epochs with early stopping (patience 30 on validation MAPE). Each $(\text{grid}, \text{model}, W)$ configuration trains independently from scratch with fixed seed (42). For the main benchmark we use a fixed chronological split (70\%/15\%/15\%).

\subsubsection*{Evaluation Metrics}
\label{sec:metrics}
Our primary metric is mean squared error, reported as a normalized percentage MSE~(\%) for cross-grid comparability, alongside mean absolute percentage error MAPE~(\%):
\begin{align}
\text{MSE (\%)} &= \frac{100}{\bar{y}}\sqrt{\frac{1}{n}\sum_{i=1}^{n}(y_i-\hat{y}_i)^2}, \\
\text{MAPE (\%)} &= \frac{100}{n}\sum_{i=1}^{n}\left|\frac{y_i-\hat{y}_i}{y_i}\right|,
\end{align}
where $\bar{y}$ is the mean load of the test set. MSE~(\%) normalizes the root mean squared error by mean load, enabling direct comparison across grids of different scale.
For generation and price signals where actuals approach zero (e.g., nighttime solar output), MAPE can be unstable. We therefore report normalized MAE (nMAE) for non-load tasks:
\begin{align}
\text{nMAE} &= \frac{100}{n\,\bar{y}}\sum_{i=1}^{n}|y_i - \hat{y}_i|, \\
\bar{y} &= \frac{1}{n}\sum_{i=1}^{n}|y_i|.
\end{align}
To characterize forecast-error tails, we use signed percentage error
$e_i = 100(\hat{y}_i-y_i)/(y_i+\epsilon)$ (with $\epsilon=10^{-6}$) and report the tail percentiles
$P_{0.5}(e)$ and $P_{99.5}(e)$.

\subsubsection*{Walk-Forward Backtest}
For weather integration analysis, we use a rolling-origin walk-forward backtesting protocol over Nov 2023--Nov 2025 (details in the Supplementary Material), aligning meteorological covariates to delayed HVAC response using empirically derived thermal lag parameters.

\subsection{Load-Only Performance Across US Grids}
\label{sec:results}

Table~\ref{tab:us_iso_table2} presents the full benchmark across US operators and prediction windows; Table~\ref{tab:tail_distributions} reports the corresponding signed-error tails for all models; Table~\ref{tab:benchmark_scorecard} summarizes row-level wins and macro averages across all 30 $(\mathrm{ISO},W)$ rows.
\begin{table*}[htbp]
\centering
\caption{\textbf{US ISO benchmark on load forecasting.} Results are reported per grid operator and prediction window ($W=24,48,72,96,168$) with fixed input length ($L=240$). For each model we report MSE (\%) and MAPE (\%) (lower is better). MSE (\%) is normalized root mean squared error as a percentage of mean load. Best values per row are highlighted in \textbf{bold}. Full tail distribution metrics ($P_{0.5}/P_{99.5}$) for all models are reported separately in Table~\ref{tab:tail_distributions}.}
\label{tab:us_iso_table2}
\begin{tabular}{lcrrrrrrrrrr}
\toprule
\multicolumn{2}{c}{} & \multicolumn{4}{c}{\textbf{SSM}} & \multicolumn{4}{c}{\textbf{Transformer}} & \multicolumn{2}{c}{\textbf{RNN}} \\
\textbf{ISO} & \textbf{$W$} & \multicolumn{2}{c}{PowerMamba} & \multicolumn{2}{c}{S-Mamba} & \multicolumn{2}{c}{iTransformer} & \multicolumn{2}{c}{PatchTST} & \multicolumn{2}{c}{LSTM} \\
 &  & MSE & MAPE & MSE & MAPE & MSE & MAPE & MSE & MAPE & MSE & MAPE \\
 &  & (\%) & (\%) & (\%) & (\%) & (\%) & (\%) & (\%) & (\%) & (\%) & (\%) \\
\midrule
\multirow{5}{*}{CAISO} & 24 & 5.12 & 3.38 & 4.74 & 3.17 & 7.30 & 5.16 & \textbf{4.70} & \textbf{3.12} & 5.91 & 4.08 \\
 & 48 & 7.35 & 4.65 & 6.18 & 4.24 & 6.42 & 4.55 & \textbf{5.86} & \textbf{3.88} & 7.76 & 5.49 \\
 & 72 & 8.70 & 5.45 & 7.47 & 5.14 & 7.21 & 5.12 & \textbf{6.73} & \textbf{4.49} & 8.83 & 6.08 \\
 & 96 & 10.73 & 6.93 & 8.58 & 5.87 & 7.68 & 5.43 & \textbf{7.28} & \textbf{4.88} & 9.98 & 6.81 \\
 & 168 & 10.43 & 7.00 & 15.28 & 9.83 & \textbf{8.66} & \textbf{6.12} & 8.87 & 6.17 & 11.02 & 7.51 \\
\midrule
\multirow{5}{*}{ISO-NE} & 24 & \textbf{8.64} & \textbf{5.86} & 8.84 & 5.95 & 9.42 & 6.41 & 8.80 & 6.06 & 9.60 & 6.74 \\
 & 48 & \textbf{11.67} & \textbf{7.77} & 11.96 & 8.00 & 12.45 & 8.29 & 11.77 & 8.01 & 12.52 & 8.58 \\
 & 72 & \textbf{13.16} & \textbf{8.93} & 13.34 & 9.05 & 14.41 & 9.56 & 13.55 & 9.13 & 19.06 & 14.02 \\
 & 96 & \textbf{14.24} & 9.83 & 14.38 & 9.99 & 15.47 & 10.34 & 14.37 & \textbf{9.72} & 17.11 & 11.89 \\
 & 168 & \textbf{15.08} & 10.58 & 15.18 & 10.61 & 16.33 & 11.18 & 15.29 & \textbf{10.55} & 19.68 & 14.24 \\
\midrule
\multirow{5}{*}{MISO} & 24 & 3.50 & 2.38 & \textbf{3.46} & \textbf{2.33} & 4.10 & 2.88 & 3.56 & 2.37 & 4.37 & 2.97 \\
 & 48 & \textbf{5.18} & 3.51 & 5.65 & 3.62 & 5.44 & 3.77 & 5.22 & \textbf{3.46} & 6.20 & 4.20 \\
 & 72 & 6.47 & 4.36 & 6.78 & 4.44 & 6.46 & 4.47 & \textbf{6.32} & \textbf{4.25} & 6.98 & 4.68 \\
 & 96 & 7.54 & 5.10 & 7.33 & 5.00 & 7.16 & 4.94 & \textbf{7.10} & \textbf{4.83} & 8.83 & 6.13 \\
 & 168 & 8.44 & 6.05 & 8.88 & 6.37 & 8.53 & 5.98 & \textbf{8.25} & \textbf{5.66} & 9.19 & 6.28 \\
\midrule
\multirow{5}{*}{PJM} & 24 & 4.72 & 3.14 & \textbf{4.51} & \textbf{2.97} & 5.16 & 3.46 & 5.00 & 3.21 & 5.47 & 3.59 \\
 & 48 & 6.59 & 4.33 & \textbf{6.29} & \textbf{4.19} & 6.94 & 4.60 & 7.19 & 4.62 & 6.94 & 4.73 \\
 & 72 & 8.31 & 5.36 & \textbf{7.84} & \textbf{5.32} & 8.22 & 5.44 & 8.78 & 5.73 & 9.08 & 6.00 \\
 & 96 & \textbf{8.59} & \textbf{5.73} & 8.88 & 6.08 & 9.04 & 6.00 & 9.64 & 6.36 & 16.09 & 10.66 \\
 & 168 & 10.91 & \textbf{6.93} & 10.47 & 7.45 & \textbf{10.37} & 7.00 & 10.41 & 6.99 & 16.39 & 11.12 \\
\midrule
\multirow{5}{*}{ERCOT} & 24 & 4.76 & 2.98 & 4.92 & 3.17 & 7.47 & 5.26 & \textbf{4.65} & \textbf{2.85} & 5.24 & 3.40 \\
 & 48 & 6.13 & 4.00 & 6.17 & 4.01 & 8.45 & 5.95 & \textbf{6.07} & \textbf{3.87} & 6.54 & 4.23 \\
 & 72 & 6.79 & 4.39 & 6.96 & 4.60 & 9.02 & 6.32 & \textbf{6.78} & \textbf{4.38} & 7.19 & 4.68 \\
 & 96 & 7.39 & 4.88 & 7.47 & 5.01 & 7.67 & 5.07 & \textbf{7.21} & \textbf{4.71} & 7.79 & 5.17 \\
 & 168 & 8.38 & 5.70 & 8.27 & 5.69 & 8.51 & 5.72 & \textbf{7.74} & \textbf{5.13} & 8.70 & 5.98 \\
\midrule
\multirow{5}{*}{NYISO} & 24 & 8.19 & 4.56 & \textbf{6.61} & \textbf{4.25} & 7.07 & 4.80 & 6.89 & 4.36 & 7.15 & 4.69 \\
 & 48 & 9.51 & 5.95 & \textbf{9.10} & \textbf{5.70} & 9.47 & 6.28 & 9.98 & 6.22 & 9.90 & 6.37 \\
 & 72 & 10.38 & \textbf{6.66} & \textbf{10.13} & 6.67 & 11.06 & 7.28 & 11.52 & 7.17 & 11.39 & 7.58 \\
 & 96 & 11.54 & 7.29 & \textbf{10.91} & \textbf{7.26} & 12.03 & 7.98 & 11.90 & 7.47 & 12.02 & 8.03 \\
 & 168 & 12.15 & \textbf{7.97} & \textbf{12.01} & 8.05 & 13.24 & 8.86 & 12.39 & 8.05 & 13.60 & 8.92 \\
\bottomrule
\end{tabular}
\end{table*}

\begin{table}[htbp]
\centering
\caption{\textbf{Tail distributions of forecast errors.} Signed percentage-error tails $P_{0.5}/P_{99.5}$ for each model across grid operators and prediction windows. Negative $P_{0.5}$ indicates under-prediction; positive $P_{99.5}$ indicates over-prediction. PwrM.: PowerMamba; iTrnsf.: iTransformer; PTST: PatchTST.}
\label{tab:tail_distributions}
\footnotesize
\setlength{\tabcolsep}{1.5pt}
\begin{tabular}{llrrrrr}
\toprule
\textbf{Grid} & \textbf{$W$} & \textbf{PwrM.} & \textbf{S-Mamba} & \textbf{iTrnsf.} & \textbf{PTST} & \textbf{LSTM} \\
 & & \multicolumn{1}{c}{(\%)} & \multicolumn{1}{c}{(\%)} & \multicolumn{1}{c}{(\%)} & \multicolumn{1}{c}{(\%)} & \multicolumn{1}{c}{(\%)} \\
\midrule
\multirow{5}{*}{CAISO} & 24 & -14.1/13.4 & -12.2/12.5 & -17.1/19.8 & \textbf{-12.6/11.6} & -15.9/14.2 \\
 & 48 & -17.1/24.9 & -14.3/18.1 & -17.9/17.6 & \textbf{-15.3/15.8} & -19.6/18.5 \\
 & 72 & -21.0/29.7 & -16.3/21.2 & -19.5/19.7 & \textbf{-17.7/16.4} & -21.6/19.4 \\
 & 96 & -20.7/37.4 & -18.5/25.1 & -19.8/21.1 & \textbf{-18.4/17.8} & -23.3/20.9 \\
 & 168 & -21.4/29.5 & -22.4/51.2 & -20.9/23.1 & \textbf{-22.1/19.8} & -23.0/33.3 \\
\midrule
\multirow{5}{*}{ISO-NE} & 24 & -18.7/34.3 & -17.5/36.2 & -21.0/35.7 & \textbf{-20.0/32.2} & -20.3/36.9 \\
 & 48 & -25.9/43.8 & -24.1/50.5 & -27.9/50.4 & \textbf{-26.8/41.5} & -27.6/45.4 \\
 & 72 & \textbf{-26.8/51.0} & -28.7/49.4 & -32.3/57.9 & -30.5/48.2 & -40.0/48.0 \\
 & 96 & -28.5/54.0 & \textbf{-29.8/52.1} & -33.6/59.6 & -31.0/51.5 & -38.4/57.2 \\
 & 168 & -32.3/52.7 & \textbf{-33.6/48.1} & -34.6/57.7 & -36.5/49.8 & -41.5/46.7 \\
\midrule
\multirow{5}{*}{MISO} & 24 & -8.5/10.8 & \textbf{-8.7/10.6} & -10.3/11.9 & -9.6/11.0 & -10.9/12.4 \\
 & 48 & \textbf{-13.2/15.1} & -14.2/15.6 & -13.7/16.4 & -13.3/16.7 & -16.6/16.0 \\
 & 72 & \textbf{-16.4/17.9} & -17.5/19.6 & -16.5/19.9 & -14.8/19.9 & -19.2/17.3 \\
 & 96 & -19.1/19.8 & -17.6/21.2 & -18.2/21.9 & \textbf{-17.0/21.4} & -19.3/21.5 \\
 & 168 & -18.0/25.2 & -19.1/25.3 & -19.1/25.7 & -19.3/23.7 & \textbf{-21.5/21.1} \\
\midrule
\multirow{5}{*}{PJM} & 24 & -11.8/13.0 & \textbf{-11.2/12.7} & -12.7/15.2 & -12.5/15.1 & -13.7/14.2 \\
 & 48 & -17.0/16.7 & \textbf{-15.7/17.7} & -16.2/21.8 & -16.9/23.2 & -16.5/17.3 \\
 & 72 & -22.0/20.9 & \textbf{-17.5/22.7} & -18.3/26.9 & -18.6/31.1 & -21.4/20.3 \\
 & 96 & \textbf{-19.6/24.5} & -19.2/25.2 & -19.5/31.6 & -21.9/33.6 & -36.7/30.8 \\
 & 168 & -30.0/21.9 & \textbf{-21.7/29.4} & -23.1/32.5 & -25.6/32.1 & -34.0/34.8 \\
\midrule
\multirow{5}{*}{ERCOT} & 24 & \textbf{-11.3/16.8} & -11.0/17.2 & -15.7/19.1 & -12.0/16.2 & -13.0/17.9 \\
 & 48 & \textbf{-13.6/20.2} & -13.7/20.8 & -17.7/22.2 & -13.3/20.7 & -16.0/20.9 \\
 & 72 & -14.8/22.9 & -14.9/23.4 & -19.3/23.9 & \textbf{-15.1/22.1} & -16.9/22.5 \\
 & 96 & \textbf{-16.4/22.6} & -15.9/23.5 & -17.8/24.9 & -16.7/22.8 & -18.2/23.3 \\
 & 168 & \textbf{-17.6/25.0} & -15.9/28.2 & -19.5/27.0 & -17.5/26.1 & -16.9/27.3 \\
\midrule
\multirow{5}{*}{NYISO} & 24 & -27.4/27.4 & \textbf{-13.9/24.6} & -15.1/27.5 & -16.0/27.3 & -15.2/25.6 \\
 & 48 & -21.6/40.3 & -20.4/37.3 & -20.8/37.5 & -23.2/37.1 & \textbf{-22.7/34.4} \\
 & 72 & -23.8/39.5 & \textbf{-22.3/37.6} & -24.5/43.2 & -27.2/40.1 & -26.0/36.3 \\
 & 96 & -30.6/39.7 & \textbf{-25.4/37.5} & -26.5/44.9 & -28.0/40.1 & -27.7/37.3 \\
 & 168 & \textbf{-31.3/34.2} & -29.8/37.1 & -29.9/43.2 & -30.4/36.1 & -34.1/33.8 \\
\bottomrule
\end{tabular}
\end{table}

\begin{table}[t]
\centering
\caption{\textbf{Benchmark scorecard.} MAPE (\%) and MSE (\%) averaged across the 30 $(\mathrm{ISO},W)$ rows. Row-level wins counted among the five main architectures.}
\label{tab:benchmark_scorecard}
\footnotesize
\setlength{\tabcolsep}{2pt}
\begin{tabular}{cccccc}
\toprule
\textbf{Model} & \textbf{Params} & \textbf{MAPE (\%)} & \textbf{MSE (\%)} & \textbf{MSE Wins} & \textbf{MAPE Wins} \\
\midrule
PowerMamba & 4.0M & 5.72 & 7.62 & 7/30 & 7/30 \\
S-Mamba & 2.0M & 5.80 & 7.58 & 9/30 & 7/30 \\
iTransformer & 6.5M & 6.14 & 7.95 & 2/30 & 1/30 \\
PatchTST & 2.0M & \textbf{5.59} & \textbf{7.53} & 12/30 & 15/30 \\
LSTM & 2.6M & 6.83 & 9.13 & 0/30 & 0/30 \\
\bottomrule
\end{tabular}
\end{table}

PatchTST emerges as the strongest load-only architecture, outperforming on the majority of evaluations and achieving the lowest average MAPE across all grid--horizon combinations (Table~\ref{tab:benchmark_scorecard}). SSMs are competitive, with particular strength on ISO-NE, PJM, and NYISO. Their $O(n)$ inference complexity makes them attractive for operational deployment. In contrast, PatchTST leads on CAISO, MISO, and ERCOT at $W=48$, suggesting that grid-specific characteristics---discussed further in Section~\ref{sec:grid_patterns}---interact with architectural inductive biases.

As expected from Section~\ref{sec:methodology}, iTransformer's cross-variate attention degenerates to near-identity with a single load variate, yielding the second-highest average error. This limitation establishes the baseline for the weather experiments in Section~\ref{sec:weather}, where additional variates restore the attention mechanism. LSTM, despite 2.6M parameters, achieves the highest average MAPE, degrading markedly at longer horizons.

For reference, published operational day-ahead MAPE values are PJM ${\sim}1.9\%$ \cite{pjm_forecast_2024}, ISO-NE ${\sim}2.1\%$ \cite{isone_forecast_2025}, and ERCOT ${\sim}2.7\%$ \cite{ercot_mtlf_2021}. Our best 24-hour results approach these values for ERCOT but remain above PJM and ISO-NE targets. This performance gap reflects factors beyond architecture---weather forecasts, ensemble methods, human analyst correction, and more frequent retraining---that our controlled benchmark intentionally excludes to isolate architectural contribution.

\subsection{Weather Integration Across Architectures and Grids}
\label{sec:weather}

We evaluate weather-integrated variants of all five architectures (Section~\ref{sec:weather_arch}) across seven US grids at $W=24$, extending the benchmark set with SWPP. Unlike the fixed-split protocol used for the main benchmark (Section~\ref{sec:results}), weather experiments use the walk-forward backtest (described in the Supplementary Material) to approximate operational conditions; consequently, absolute MAPE values are not directly comparable between the two sections, though within-section deltas are valid. Each model uses its architecture-specific weather fusion strategy with thermal-lag alignment. \textit{Important assumption:} weather covariates use \textit{observed} (reanalysis) meteorological data rather than numerical weather prediction (NWP) forecasts, isolating neural-architecture differences from NWP forecast skill. Operational deployment would introduce additional error from imperfect weather forecasts, particularly at longer horizons. Three conditions are tested: \textit{none} (load only), \textit{temporal} (load + temporal embeddings as a matched-token control), and \textit{weather} (load + weather covariates + temporal embeddings). This yields 105 experiments ($7 \times 5 \times 3$).

Table~\ref{tab:weather_results} reports baseline and weather-integrated MAPE for all model--grid combinations.

\begin{table*}[htbp]
\centering
\caption{\textbf{Weather integration results: Load-only vs.\ Weather-integrated MAPE (\%) across 7 US grids ($W=24$).} Bold marks the best MAPE per grid within each condition. `Load' uses historical load only; `Weather' adds meteorological covariates. All models use architecture-specific weather fusion (Section~\ref{sec:weather_arch}).}
\label{tab:weather_results}
\footnotesize
\setlength{\tabcolsep}{3pt}
\begin{tabular}{l r c r c r c r c r c}
\toprule
 & \multicolumn{2}{c}{iTransformer} & \multicolumn{2}{c}{PatchTST} & \multicolumn{2}{c}{PowerMamba} & \multicolumn{2}{c}{S-Mamba} & \multicolumn{2}{c}{LSTM} \\
\cmidrule(lr){2-3} \cmidrule(lr){4-5} \cmidrule(lr){6-7} \cmidrule(lr){8-9} \cmidrule(lr){10-11}
\textbf{Grid} & Load & Weather & Load & Weather & Load & Weather & Load & Weather & Load & Weather \\
\midrule
CAISO & 5.16 & 3.19 & \textbf{3.12} & 3.20 & 3.38 & 3.51 & 3.17 & 3.62 & 4.08 & 3.71 \\
ISO-NE & 6.41 & 5.01 & 6.06 & 4.41 & \textbf{5.86} & \textbf{3.62} & 5.95 & 3.63 & 6.74 & 6.20 \\
MISO & 2.88 & 2.25 & 2.37 & 2.31 & 2.38 & \textbf{2.11} & \textbf{2.33} & 2.16 & 2.97 & 2.62 \\
PJM & 3.46 & \textbf{2.73} & 3.21 & 3.04 & 3.14 & 2.78 & \textbf{2.97} & \textbf{2.73} & 3.59 & 3.33 \\
SWPP & 5.94 & 3.32 & 3.47 & 3.21 & 3.50 & 3.18 & \textbf{3.43} & \textbf{3.11} & 4.37 & 3.70 \\
ERCOT & 5.26 & 2.39 & \textbf{2.85} & 2.05 & 2.98 & 2.09 & 3.17 & \textbf{1.96} & 3.40 & 3.37 \\
NYISO & 4.80 & 3.69 & 4.36 & 3.56 & 4.56 & 3.30 & \textbf{4.25} & \textbf{3.11} & 4.69 & 4.36 \\
\bottomrule
\end{tabular}
\end{table*}

\noindent Table~\ref{tab:weather_delta} summarizes the weather integration impact as $\Delta$MAPE (weather $-$ none, in percentage points).

\begin{table}[htbp]
\centering
\caption{\textbf{Weather integration impact ($\Delta$MAPE, percentage points).} Negative values indicate improvement with weather covariates. Bold marks the largest improvement per grid.}
\label{tab:weather_delta}
\footnotesize
\setlength{\tabcolsep}{4pt}
\begin{tabular}{lrrrrr}
\toprule
\textbf{Grid} & iTransformer & PatchTST & PowerMamba & S-Mamba & LSTM \\
\midrule
CAISO & \textbf{-1.97} & +0.08 & +0.13 & +0.45 & -0.37 \\
ISO-NE & -1.40 & -1.65 & -2.24 & \textbf{-2.32} & -0.54 \\
MISO & \textbf{-0.63} & -0.06 & -0.27 & -0.17 & -0.35 \\
PJM & \textbf{-0.73} & -0.17 & -0.36 & -0.24 & -0.26 \\
SWPP & \textbf{-2.62} & -0.26 & -0.32 & -0.32 & -0.67 \\
ERCOT & \textbf{-2.87} & -0.80 & -0.89 & -1.21 & -0.03 \\
NYISO & -1.11 & -0.80 & \textbf{-1.26} & -1.14 & -0.33 \\
\midrule
\textit{Average} & \textbf{-1.62} & -0.52 & -0.74 & -0.71 & -0.36 \\
\bottomrule
\end{tabular}
\end{table}

Weather covariates improve all five architectures on the majority of grids, but the magnitude of improvement varies substantially by architecture (Table~\ref{tab:weather_delta}). iTransformer benefits significantly more than PatchTST, with the gap widening on weather-sensitive grids (ERCOT and SWPP).

This benefit gap is partly explained by iTransformer's degraded single-variate baseline: adding weather variates restores the multi-token attention that the architecture requires to function as designed. PatchTST's channel-independent baseline is already strong, leaving less room for improvement. A temporal-embedding control experiment supports this interpretation. Temporal-only tokens improve iTransformer and LSTM, but have negligible effect on PatchTST, PowerMamba, and S-Mamba. This confirms that iTransformer benefits from any additional tokens, not specifically from weather content.

With weather features included, SSMs become the strongest models overall. S-Mamba achieves the best MAPE on SWPP, ERCOT, and NYISO, while PowerMamba leads on ISO-NE and MISO. iTransformer outperforms on CAISO and ties PJM.

\subsubsection*{Capacity-Controlled Comparison}
\label{sec:fair_comparison}

The weather benefit gap between iTransformer ($-1.62$ percentage points) and PatchTST ($-0.52$ percentage points) could reflect parameter count differences rather than architectural advantages: iTransformer uses $d_\text{model}=512$ (6.5M parameters) while PatchTST uses $d_\text{model}=256$ (2.0M). To disentangle capacity from architecture, we train two additional model variants. We evaluate Tier~1 ($d_\text{model}=512$), comparing the paper-faithful iTransformer (6.5M parameters) against a scaled PatchTST$_L$ ($d_\text{model}=512$, 9.84M parameters); and Tier~2 ($d_\text{model}=256$), comparing a shrunk iTransformer$_S$ (2.44M parameters) against the paper-faithful PatchTST (2.0M parameters). Note that matching $d_\text{model}$ does not equalize total parameter counts because the architectures differ structurally: PatchTST's patch embedding, multi-head self-attention, and feed-forward layers scale differently with $d_\text{model}$ than iTransformer's variate-tokenization and cross-variate attention. We therefore match the hidden dimension---the primary capacity bottleneck---and report exact counts for transparency.

Each variant is evaluated across all seven grids under the same three conditions (none, temporal, weather), yielding 84 additional experiments. Table~\ref{tab:weather_fair} reports the $\Delta$MAPE (weather $-$ none) for all four variants alongside the iTransformer--PatchTST gap ($\Delta$Gap) at each tier.

\begin{table}[htbp]
\centering
\caption{\textbf{Parameter-controlled weather comparison.} MAPE (\%) for baseline (None) and weather-integrated (Wx) variants at matched $d_\text{model}$ sizes. $\Delta$ is Weather$-$None (pp).}
\label{tab:weather_fair}
\scriptsize
\setlength{\tabcolsep}{2pt}
\begin{tabular}{l ccc ccc}
\toprule
& \multicolumn{3}{c}{\textbf{Tier 1} ($d=512$)} & \multicolumn{3}{c}{\textbf{Tier 2} ($d=256$)} \\
\cmidrule(lr){2-4} \cmidrule(lr){5-7}
\textbf{Grid} & iTransformer & PatchTST$_L$ & $\Delta$Gap & iTransformer$_S$ & PatchTST & $\Delta$Gap \\
& (6.44M) & (9.84M) & & (2.44M) & (1.77M) & \\
\midrule
CAISO & \textbf{-1.97} & +0.20 & -2.17 & -0.75 & +0.08 & -0.83 \\
ISO-NE & -1.40 & -1.62 & +0.22 & -0.71 & \textbf{-1.65} & +0.94 \\
MISO & -0.63 & +0.13 & -0.76 & \textbf{-2.49} & -0.06 & -2.43 \\
PJM & -0.73 & -0.02 & -0.71 & \textbf{-2.28} & -0.17 & -2.11 \\
SWPP & \textbf{-2.62} & -0.16 & -2.46 & -2.32 & -0.26 & -2.06 \\
ERCOT & \textbf{-2.87} & -0.56 & -2.31 & -1.96 & -0.80 & -1.16 \\
NYISO & \textbf{-1.11} & -0.67 & -0.44 & -0.72 & -0.80 & +0.08 \\
\bottomrule
\end{tabular}
\end{table}

The results confirm that the weather advantage is architectural, not capacity-driven. Scaling PatchTST to $d=512$ degrades its average $\Delta$MAPE, indicating that additional capacity does not improve weather utilization in a channel-independent design. Conversely, shrinking iTransformer to $d=256$ preserves its weather benefit, demonstrating robustness to a parameter reduction. The gap is consistent across both tiers: iTransformer outperforms its capacity-matched PatchTST counterpart on the majority of grids at each tier (Table~\ref{tab:weather_fair}). ISO-NE is the only grid where PatchTST variants benefit more at both tiers.

\subsubsection*{Grid-Level Determinants of Weather Benefit}
\label{sec:weather_predictors}

The preceding sections establish that weather features help all architectures and that the benefit magnitude is architecturally determined. A natural follow-up question is which \textit{grids} benefit the most. The na\"ive hypothesis---that grids in extreme climates should benefit more---would suggest that temperature variability ($\sigma_T$) predicts weather integration value. We examine this across the seven US grids by comparing each grid's average $\Delta$MAPE with $\sigma_T$ and baseline forecast difficulty.

\begin{figure*}[htbp]
\centering
\includegraphics[width=0.95\textwidth]{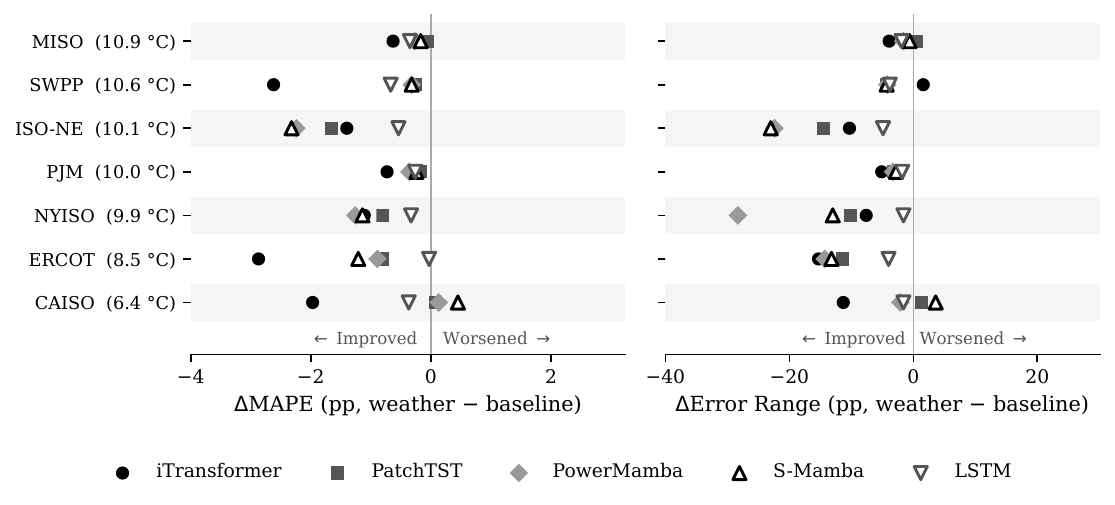}
\caption{\textbf{Weather integration improves accuracy and narrows the error range, with stronger effects on weather-sensitive grids.} The plot illustrates the change in MAPE (left) and Error Range ($P_{99.5} - P_{0.5}$, right) for five neural architectures when transitioning from load-only to weather-integrated forecasting ($W=24$). Negative values indicate improvement. Grids are ordered by descending temperature variability ($\sigma_T$). High-variability grids (e.g., ISO-NE, ERCOT) exhibit substantially larger improvements in both mean accuracy and tail extremes compared to mild grids (e.g., CAISO).}
\label{fig:weather_benefit}
\end{figure*}

Figure~\ref{fig:weather_benefit} shows that temperature variability is necessary but not sufficient for weather benefit. Grids with low $\sigma_T$ (such as CAISO) gain little from weather features, while high-variability grids show larger but heterogeneous benefits. The interaction between climate variability and baseline difficulty determines the realized benefit.

\textit{Low variability, low benefit.} CAISO shows near-zero or positive $\Delta$MAPE across all five models, consistent with its mild Mediterranean climate where load patterns are highly predictable from temporal features alone.

\textit{High variability, high benefit.} ISO-NE and ERCOT exhibit the largest weather benefits, where heating- and cooling-driven demand creates weather-dependent variance that temporal patterns cannot capture.

\textit{High variability, modest benefit.} MISO has the highest temperature variability but the smallest weather benefit among the high-variability grids. Its low baseline MAPE indicates that spatial smoothing across its large footprint already reduces weather-driven variance, leaving little residual signal for weather covariates to exploit.

These patterns suggest a ceiling effect in which temperature variability determines the \textit{potential} for weather-driven load variance, but baseline forecast difficulty determines how much of that variance remains unexploited. When load-only models already achieve MAPE near 2\% (MISO, PJM), temporal patterns capture most variance and weather covariates provide diminishing returns. Weather integration is most valuable where high climate variability coincides with high baseline error---typically electrically isolated grids (ERCOT), smaller systems (SWPP), and weather-sensitive regions (ISO-NE).

\subsection{Generalization Beyond Load}
\label{sec:multitype}

To test whether architectural rankings generalize beyond load, we extend the benchmark to solar generation (6 core ISOs), wind generation (7 grids: the 6 core ISOs plus SWPP), wholesale electricity price (4 grids: CAISO, ERCOT, NYISO, ISO-NE), and ERCOT ancillary services (4 types: RegUp, RegDown, RRS, NonSpin)---totaling 21 grid--signal combinations, 5 models, and 2 conditions (none, weather) at $W=24$. Because MAPE diverges when actuals approach zero (e.g., nighttime solar), we use nMAE as the primary metric for all non-load tasks.

\begin{table}[htbp]
\centering
\caption{\textbf{Multitype forecasting scorecard: average nMAE (\%) by signal type and model (no weather, $W=24$).} Bold marks the best model per row. Win counts reflect per-grid nMAE victories.\textsuperscript{$\dagger$}CAISO partial: 14 EIA-CISO logs lack nMAE (older metrics code).}
\label{tab:multitype_scorecard}
\scriptsize
\setlength{\tabcolsep}{3pt}
\begin{tabular}{lccccc}
\toprule
\textbf{Signal Type} & PatchTST & S-Mamba & PowerMamba & iTransformer & LSTM \\
\midrule
Solar & \textbf{23.1} & 31.8 & 30.6 & 68.8 & 38.5 \\
Wind & 36.2 & \textbf{35.4} & 35.7 & 36.4 & 37.2 \\
Price & 30.9 & \textbf{27.5} & 34.1 & 60.4 & 37.5 \\
Ancillary Svc. & \textbf{79.3} & 119.9 & 139.1 & 597.2 & 131.5 \\
\midrule
\textit{nMAE Wins} & 11/18 & 5/18 & 2/18 & 0/18 & 0/18 \\
\bottomrule
\end{tabular}
\end{table}

Table~\ref{tab:multitype_scorecard} reveals that signal structure---rather than model family---determines the architectural ranking. PatchTST dominates signals with strong diurnal periodicity: solar and ancillary services, where its fixed-length patching aligns with 24-hour cycles. S-Mamba leads on signals with weaker periodic structure: wind and price, where continuous state dynamics better capture aperiodic dependencies. iTransformer's near-identity limitation (Section~\ref{sec:results}) generalizes across all signal types, yielding poor performance on single-variate tasks.

Weather benefit also varies by task type in ways that differ from the load pattern. For wind, SSMs benefit most, reversing the load ranking where iTransformer gains the most. For price, iTransformer benefits strongly, recovering from its weak single-variate baseline. Solar shows mixed results: PatchTST and iTransformer improve while SSMs slightly degrade.

\section{Discussion}
\label{sec:discussion}

\subsection{Channel-Independent vs.\ Cross-Variate Attention}

The load-only vs.\ weather-augmented ranking reversal traces to a fundamental architectural difference. iTransformer's symmetric variate tokenization treats weather features as first-class tokens in the same attention space as load, enabling fine-grained load--weather interaction at each time step. PatchTST processes channels independently, so weather information can only influence load prediction indirectly through shared representations. This division explains the $3\times$ weather benefit gap (Table~\ref{tab:weather_delta}) and why it widens on weather-sensitive grids where hour-by-hour weather dynamics drive load variation. The capacity-controlled experiments (Section~\ref{sec:fair_comparison}) confirm this is architectural, not capacity-driven.

For practitioners, this suggests deploying PatchTST or SSMs for load-only forecasting, and switching to iTransformer or weather-aware SSM variants when weather forecasts are available.

\subsection{Grid-Specific Performance Patterns}
\label{sec:grid_patterns}

The benchmark reveals a systematic split: PatchTST leads on CAISO, MISO, and ERCOT, while SSMs lead on ISO-NE, PJM, and NYISO. We attribute this to the interaction between architectural inductive biases and load-profile regularity.

PatchTST's fixed-length patches ($P=16$ hours) align well with grids dominated by strong, regular diurnal patterns. CAISO's duck curve creates a stereotyped daily shape; MISO and ERCOT, as large summer-peaking systems, exhibit similarly predictable daily peaks. Patching converts these regular structures into a small number of informative tokens. In contrast, PJM and NYISO serve dense, diverse load bases where daily patterns are modulated by irregular factors---subway schedules, commercial building occupancy, and diverse industrial loads. SSMs' continuous state dynamics adapt to these irregularities through selective gating, whereas PatchTST's fixed patch boundaries may straddle meaningful transitions. ISO-NE's winter-peaking profile, driven by electric heating with weather-dependent intensity, similarly rewards the adaptive processing that SSMs provide.

This interpretation is consistent with the multitype results (Section~\ref{sec:multitype}), where PatchTST leads on signals with strong diurnal periodicity (solar) and SSMs lead on less periodic signals (wind, price).

\subsection{Implications for Weather Integration}
\label{sec:weather_discussion}

Three practical implications emerge for operators:

\textit{Prioritize by baseline difficulty, not climate extremity.} Operators should invest in weather integration on grids where load-only forecasts are weakest (ERCOT, ISO-NE), not where temperature swings are largest (MISO). Geographic load diversity in large interconnections naturally smooths weather shocks, creating a ceiling effect that reduces the marginal value of explicit weather features.

\textit{Match architecture to input availability.} iTransformer's single-variate weakness (avg 4.84\% baseline) makes it unsuitable for load-only deployment, but its cross-variate attention becomes the strongest weather utilization mechanism when covariates are available. This input-dependent ranking means operators should maintain different model configurations for different data-availability scenarios.

\textit{Weather-aware SSMs combine accuracy with efficiency.} S-Mamba and PowerMamba achieve the best absolute MAPE on 5/7 grids with weather features while retaining $O(n)$ inference complexity---making them attractive for real-time operational deployment where both accuracy and latency matter.

\subsection{Limitations}

\textit{Statistical robustness.} Results use a fixed random seed (42). While the large experiment count (30 grid--horizon pairs per model) provides a robust aggregate view, multi-seed evaluation would quantify per-grid variance. We also do not perform formal significance tests (e.g., paired Wilcoxon signed-rank) on per-timestamp errors; formal testing would strengthen conclusions for rows with narrow margins.

\textit{Benchmark coverage.} The weather benchmark covers $W=24$ only; extension to longer horizons is planned. We do not include foundation models (Chronos \cite{ansari2024chronos}, TimesFM \cite{das2024timesfm}) or recent architectures (TimeMixer \cite{wang2024timemixer}, hybrid Mamba-Transformer models). The multitype benchmark (Section~\ref{sec:multitype}) should be considered preliminary for ancillary services, where only ERCOT data is available.

\textit{Error decomposition and latency.} We report aggregate MAPE and tail percentiles but do not stratify errors by peak vs.\ off-peak periods, which would provide more operationally actionable guidance. We also use parameter count rather than wall-clock inference time as the efficiency metric; actual latency measurements would better characterize the $O(n)$ vs.\ $O(n^2)$ distinction.

\textit{Context length.} We fix $L=240$ hours following PowerMamba \cite{menati2024powermamba} but do not ablate this choice. Future work should evaluate the accuracy--context trade-off.

\section{Conclusion}

We benchmarked five models---two SSMs, two Transformers, and an LSTM---across six US ISOs, with weather integration on seven grids and multitype generalization to solar, wind, price, and ancillary-service forecasting. This systematic evaluation yielded three key findings.

First, no single architecture dominates; input availability determines the optimal choice. PatchTST and SSMs lead for load-only forecasting, but this ranking reverses when weather covariates are available: cross-variate attention (iTransformer) and weather-aware SSMs become strongest. Parameter-controlled experiments confirm this reversal is architectural, not capacity-driven. Operators should select architectures based on what inputs their pipeline provides.

Second, weather integration value is predictable from grid characteristics. Baseline forecast difficulty---not climate extremity---determines how much weather features help. Isolated and smaller grids benefit most; large interconnections with geographic load diversity show ceiling effects. This provides a simple heuristic for prioritizing weather integration investments.

Third, architectural rankings are task-dependent. PatchTST's fixed-length patching excels on signals with strong diurnal periodicity (load on regular grids, solar), while SSMs' adaptive state dynamics lead on less periodic signals (wind, price) and grids with irregular load patterns. This task--architecture interaction means that benchmarks on a single signal type cannot predict performance on others.

These findings provide actionable, input-dependent deployment guidance. Beyond specific recommendations, the benchmark establishes a reproducible, multi-ISO evaluation framework that future work can extend with foundation models, longer forecast horizons, and richer exogenous features. Our configurations, trained checkpoints, and evaluation code are publicly available at \url{https://github.com/gramm-ai/grid-forecast-benchmark}.

\section*{Data Availability}
Hourly load data are publicly available from the U.S. Energy Information Administration via EIA-930 \cite{eia930}. Weather covariates were obtained from Open-Meteo hourly weather products. All preprocessing code, model configurations, trained checkpoints, and evaluation scripts are available at \url{https://github.com/gramm-ai/grid-forecast-benchmark}.


\end{document}